\documentclass[letterpaper, 10 pt, conference]{ieeeconf}
\IEEEoverridecommandlockouts
\usepackage{cite}
\usepackage{amsmath,amssymb,amsfonts}
\usepackage{algorithmic}
\usepackage{graphicx}
\usepackage{cuted}
\usepackage{capt-of}
\usepackage{textcomp}
\usepackage{xcolor}
\usepackage{booktabs}
\usepackage{tikz}
\usepackage{url}
\usepackage{multirow}
\usetikzlibrary{arrows.meta,calc,positioning}
\def\BibTeX{{\rm B\kern-.05em{\sc i\kern-.025em b}\kern-.08em
    T\kern-.1667em\lower.7ex\hbox{E}\kern-.125emX}}
\newcommand{\sysname}{4DSynth}

\title{\LARGE \bf
\sysname: Controllable Procedural World Synthesis for Dynamic Embodied Simulation
}

\author{%
\authorblockN{Zehao Qi$^{1,*}$, Haochen Luo$^{2,*}$, Jia-Wang Bian$^{1}$, Zeyu Ma$^{3}$, and Shuyang Sun$^{4}$}
\authorblockA{\small
$^{1}$Nanyang Technological University, Singapore \quad
$^{2}$University of Oxford, UK\\
$^{3}$Princeton University, USA \quad
$^{4}$Google DeepMind, USA\\
$^{*}$Equal contribution\\
{\tt \{zehao.qi.mail,jiawang.bian\}@gmail.com} \quad
{\tt haochen.luo@outlook.com}\\
{\tt zeyum@princeton.edu} \quad
}
}

\begin{document}

\maketitle
\thispagestyle{empty}
\pagestyle{empty}

\begin{strip}
  \centering
  \includegraphics[width=0.99\textwidth]{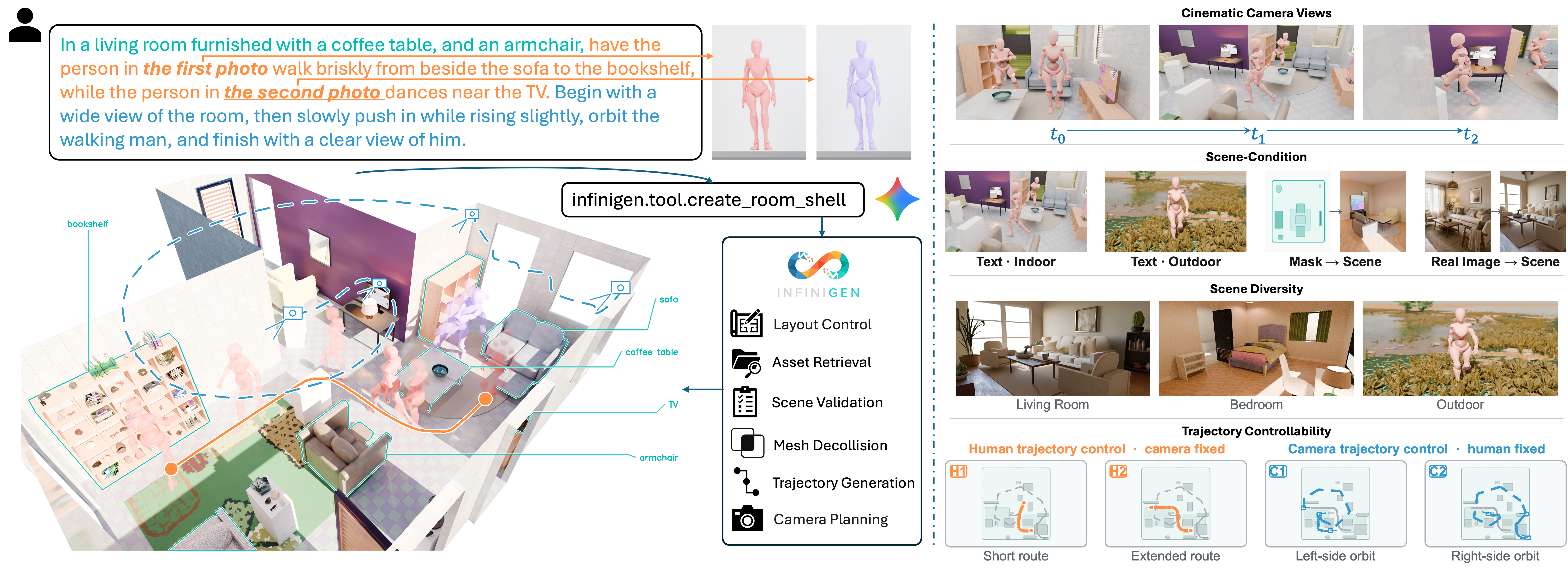}
  \captionof{figure}{System overview. A natural-language director instruction
  specifies scene, actor, and camera intent. \sysname{} builds an editable
  scene through native indoor or outdoor generation, layout-conditioned
  synthesis from text or masks, or single-image real-to-sim compilation, then
  places animated actors and plans actor and camera trajectories on the
  finished geometry. The panels on the right show control over camera views,
  scene conditions, scene types, and actor--camera trajectories.}
  \label{fig:teaser}
\end{strip}

\begin{abstract}
Embodied agents need environments that are visually diverse, physically
interactive, and changing over time. Procedural simulators can generate large
interactive scene collections, and recent 4D generators produce compelling
visual dynamics. Combining these properties in one environment, however, still
demands extensive manual effort, and the result is rarely editable or
controllable enough to reuse at scale.

We present \sysname{}, a controllable procedural system that turns a
natural-language description, a blueprint mask, or a single photograph into an
editable 4D environment with explicit geometry, animated actors, collision-free
trajectories, and physics-ready simulation state. Multiple scene routes share
one geometry-grounded representation, so the same pipeline handles animation,
camera planning, rendering, and task generation.

To validate the full pipeline, we construct 4DSynth-Nav, an interactive
navigation benchmark generated entirely from \sysname{}'s procedural scenes.
Two vision-language models evaluated across three difficulty tiers both fail
the majority of tasks and stall after early subtasks. The same procedural
controllability that produces these environments also makes each failure
reproducible and each difficulty axis independently tunable. Together, this paper presents both a generation pipeline and the scalable benchmark, offering a practical
foundation for developing and evaluating embodied agents.\end{abstract}


\section{Introduction}

Embodied agents need environments that are visually varied, physically
structured, and changing over time. Procedural simulators can generate large
collections of interactive houses~\cite{procthor,infinigenindoors},
language-guided systems make individual scenes cheaper to
specify~\cite{holodeck}, and dynamic simulators add human motion and
human--robot interaction~\cite{habitat3}. These capabilities, however, live in
separate scene representations and separate authoring pipelines. Getting from
a user's conditions, whatever form they take, to one editable scene whose
geometry, actor motion, camera motion, and simulation state agree with each
other is still hard.

Fig.~\ref{fig:teaser} summarizes our pipeline. A natural-language director
instruction gives high-level scene, actor, and camera intent. An agentic
controller turns that intent into a scene by native procedural generation, by
layout-conditioned synthesis from text or masks, or by single-image
real-to-sim compilation, and then places animated actors and plans actor and
camera trajectories on the finished geometry. Throughout the paper we use
\emph{4D environment} to mean an explicit 3D scene plus time-indexed actor and
camera states, not a purely visual dynamic representation. Keeping the
representation explicit is what lets object identity, support relations,
collision geometry, and procedural parameters survive all the way through
rendering and simulation.

Our contributions are threefold:
\begin{itemize}
  \item A unified procedural authoring architecture in which four
  scene-realization routes feed one shared Stage and one provenance-preserving
  WorldState, instead of each route having its own downstream pipeline.
  \item Geometry-grounded 4D synthesis that measures animated assets, plans
  collision-free actor and camera trajectories on the finished scene geometry,
  and derives physics-ready OpenUSD output from the procedural scene semantics.
  \item 4DSynth-Nav, a suite of 333 automatically validated navigation and
  pick-and-place tasks with animated obstacles, which we use to pin down where
  two vision-language agents fail.
\end{itemize}

\section{Related Work}\label{sec:related}

\textbf{Embodied simulation platforms.}
AI2-THOR~\cite{ai2thor}, Habitat~\cite{habitat}, iGibson
2.0~\cite{igibson2}, and SAPIEN~\cite{sapien} support navigation, household
interaction, and robot learning in interactive simulation. Their emphasis is
on fast execution, articulated objects, and reusable task APIs, usually over
scene collections that already exist. Ours is complementary: we construct the
editable world itself---geometry, moving actors, camera motion, simulation
state, and downstream tasks---from whatever conditions the user provides.

\textbf{Procedural and language-guided environments.}
ProcTHOR samples large collections of interactive houses for embodied
learning~\cite{procthor}; Infinigen and Infinigen Indoors generate
photorealistic, fully procedural natural and indoor
scenes~\cite{infinigen,infinigenindoors}. Holodeck takes a different route and
has a language model translate open-ended prompts into relational constraints
over retrieved 3D assets~\cite{holodeck}. \sysname{} builds on procedural
generation but solves a different systems problem: bringing native generation,
authored layouts, and reconstructed rooms into one geometry-grounded pipeline
for animation and simulation.

\textbf{4D generation and real-to-sim.}
Recent text-to-4D methods produce visually compelling dynamic representations
with diffusion guidance and deformable radiance or Gaussian
fields~\cite{fourdfy,ayg}. Their output is meant for novel-view rendering;
\sysname{} instead keeps explicit objects, walkability, editable procedural
parameters, and simulator collision geometry. On the real-to-sim side, Digital
Cousins automatically builds affordance-preserving scene variants for policy
learning~\cite{digitalcousins}, and VIGA reconstructs scenes through
interleaved inverse-graphics reasoning~\cite{viga}. Our single-image route
uses reconstruction as evidence but maps every observed instance to a
procedural asset that can actually be instantiated, trading exact appearance
for editability and physics export.

\textbf{Embodied simulation with dynamic agents.}
BEHAVIOR-1K and OmniGibson offer physics-rich household
activities~\cite{behavior1k}, and Habitat~3.0 adds humanoids and collaborative
human--robot tasks~\cite{habitat3}. 4DSynth-Nav is deliberately narrower: a
navigation and pick-and-place benchmark instantiated automatically from our
generated scenes, meant to show how visual agents react to controlled changes
in task horizon, initial visibility, and animated obstacles, not to replace
broad activity or collaboration benchmarks.

\section{Method}\label{sec:method}

\begin{figure*}[t]
  \centering
  \includegraphics[width=0.99\textwidth]{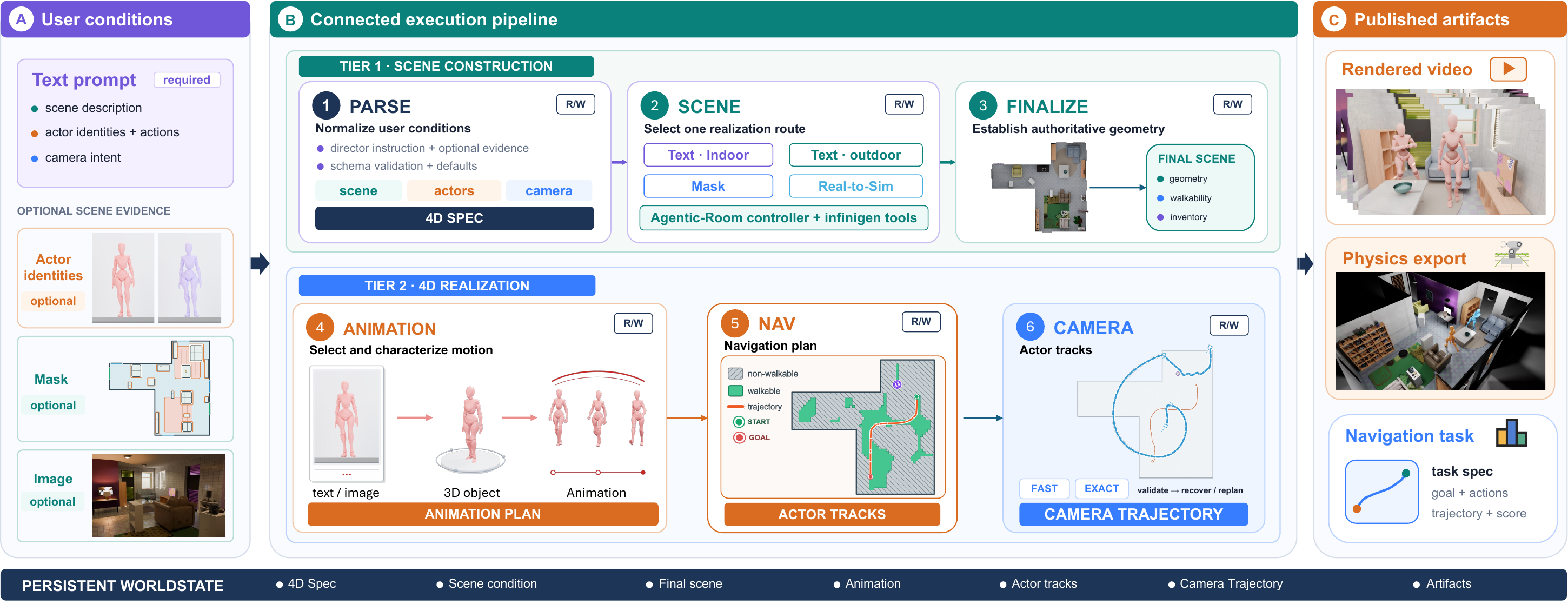}
  \caption{\sysname{} as a two-tier execution pipeline. \textsc{Parse}
  normalizes the user conditions into a 4D specification. Tier~1 realizes the
  scene through one of four routes; \textsc{Finalize} then fixes the geometry,
  walkability, and inventory of the final Stage (only native outdoor scenes go
  through coarse-to-final materialization). In parallel, \textsc{Animation}
  selects, imports, and measures motion assets to produce an animation plan;
  \textsc{Nav} joins that plan with the final Stage to produce collision-free
  actor tracks, and \textsc{Camera} plans and validates a camera trajectory. A
  persistent WorldState records the intermediate results and their provenance,
  and the completed state is used to publish rendered video, physics exports,
  and navigation tasks.}
  \label{fig:pipeline}
\end{figure*}

\sysname{} turns user conditions into an editable, animated environment
together with observations that are consistent with its geometry
(Fig.~\ref{fig:pipeline}). The architecture separates semantic control, scene
realization, and geometry-grounded 4D synthesis. A constrained language-model
interface first writes the requested scene, actors and actions, spatial goals,
and camera intent into a schema-validated 4D specification. The four
realization routes---native indoor generation, native outdoor generation,
layout-conditioned synthesis from text or masks, and single-image real-to-sim
compilation---fall into two source families: native procedural generation and
prebuilt-scene compilation.

Both families end in a \emph{Stage}, our internal record of the finished
geometry, its walkable area, and its contents; we use \emph{OpenUSD stage}
only for the exported simulator scene. Actor synthesis works on the Stage to
turn measured character assets and requested actions into collision-free
tracks, and observation synthesis realizes camera intent against the same
geometry and motion. Baking combines scene, actor tracks, and camera track into
an editable 4D environment, from which rendered video, physics simulation, and
navigation tasks are derived. A persistent WorldState keeps all of these
representations, and where each came from, across the whole pipeline.

Native outdoor scenes add one complication. Infinigen conditions detailed
terrain and population on the region the camera sees, but a useful camera
depends on where the actors can go. The outdoor branch therefore starts from a
coarse scene, uses it to establish provisional motion and viewing support,
realizes native detail inside that region, and only then extracts the final
Stage for actor and camera planning.

The rest of this section covers scene synthesis
(Section~\ref{sec:m-synthesis}), 4D animation on the shared Stage
(Section~\ref{sec:m-4d}), observation synthesis (Section~\ref{sec:m-camera}),
and physics-ready export (Section~\ref{sec:m-export}).

\subsection{Controllable Scene Synthesis}\label{sec:m-synthesis}

The 4D backbone accepts two kinds of scene source, both mapped to the same
Stage.

\textbf{Native generative source.}
Given only a category-level request, the system synthesizes the scene
procedurally: an indoor scene as a single room of the requested type or as a
whole multi-room home furnished by constraint-based
placement~\cite{infinigenindoors}; an outdoor scene as one of eight terrain
biomes (forest, desert, coast, mountain, plain, canyon, cliff, or arctic) with
vegetation and scatter assets~\cite{infinigen}. The category picks the biome
and nothing else; its continuous parameters are left to the generator. We
record the random seed so that any sampled scene can be regenerated.

\begin{figure*}[!t]
  \centering
  \includegraphics[width=\textwidth]{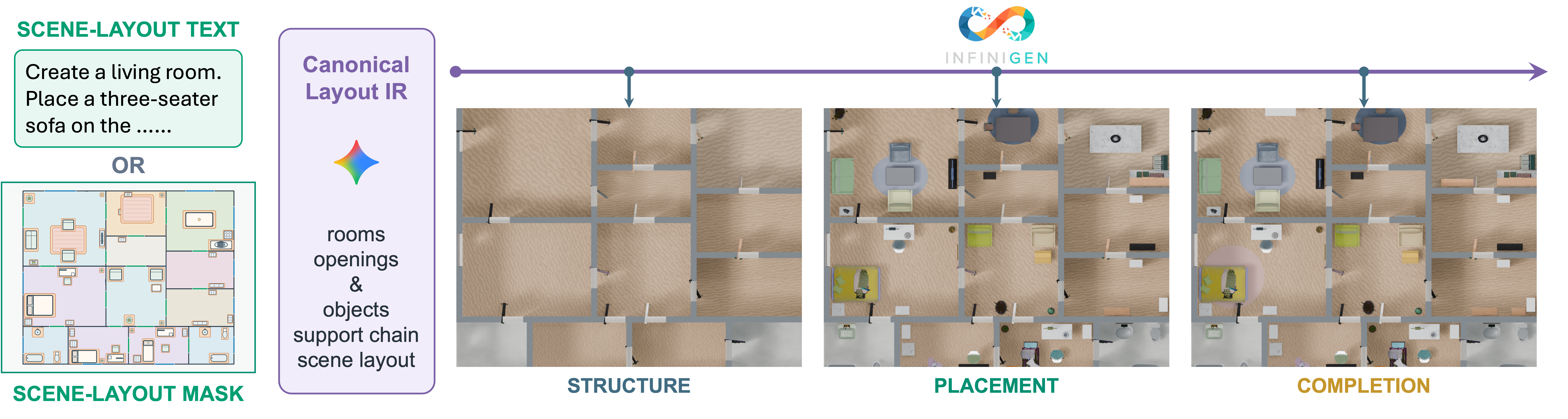}
  \caption{Text- and mask-conditioned scene compilation. A scene-layout
  description or an annotated mask is normalized into a Canonical Layout IR
  recording rooms, openings, objects, support chains, and metric layout. The
  shared procedural backend builds the room structure, instantiates and places
  the specified assets, and fills in compatible content during completion. The
  three bird's-eye-view panels show the same scene after structure
  construction, placement, and completion.}
  \label{fig:text-mask-pipeline}
\end{figure*}

\begin{figure*}[t]
  \centering
  \includegraphics[width=0.99\textwidth]{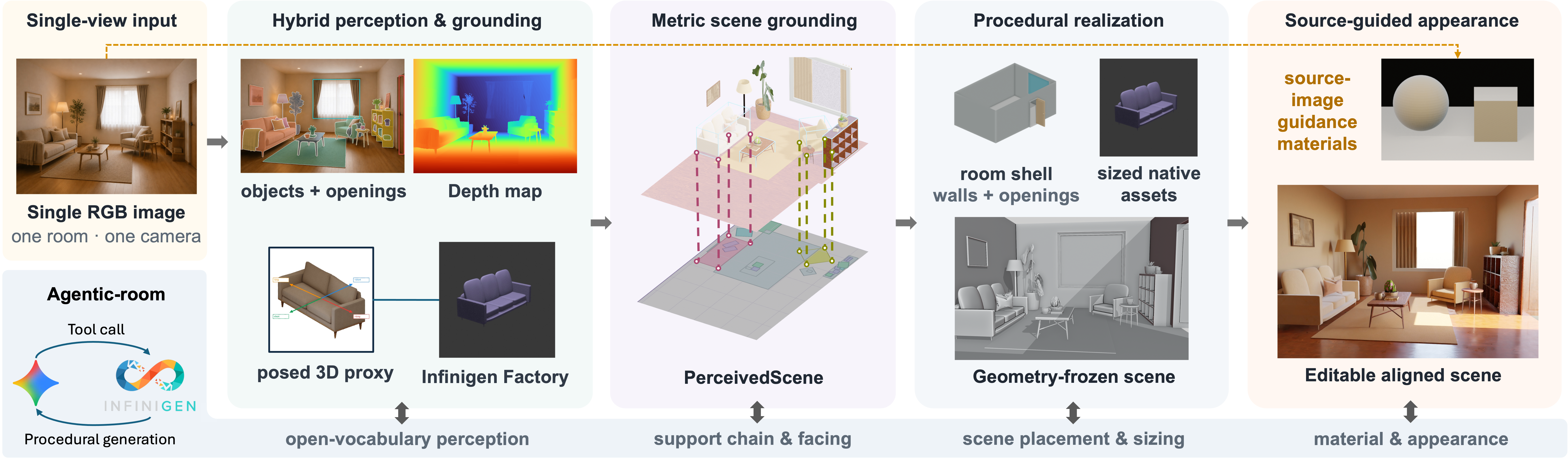}
  \caption{Single-image real-to-sim compilation. Hybrid perception extracts
  objects, openings, metric depth, and posed 3D proxies; grounding links each
  observed instance to a procedural factory. Metric reasoning consolidates this
  evidence into a PerceivedScene with room geometry, object scale, support, and
  facing. Procedural realization builds the room shell, instantiates sized
  native assets, and freezes the geometry before source-guided material
  alignment produces an editable scene that matches the input view.}
  \label{fig:r2s-pipeline}
\end{figure*}

\textbf{Prebuilt-scene source.}
Alternatively, an upstream layout compiler or real-to-sim compiler
(Figs.~\ref{fig:text-mask-pipeline} and~\ref{fig:r2s-pipeline}) hands over a
fully realized scene. The backbone leaves its geometry alone,
re-extracts and validates the Stage, and runs the same actor, navigation,
camera, and rendering stages as for native scenes. Raw layouts and photographs
are thus compiled upstream; the backbone only ever sees the realized scene.

\textbf{Text- and mask-conditioned indoor synthesis.}
Text and masks give complementary kinds of control
(Fig.~\ref{fig:text-mask-pipeline}). A scene-layout description fixes what
should be in the scene but not where, so a constrained language model proposes
rooms, openings, object footprints, orientations, and support relations. An
annotated mask already fixes the spatial layout and the object count; we keep
its authored geometry and infer only the missing semantics. Both inputs are
normalized into a Canonical Layout intermediate representation (IR) of rooms,
openings, oriented object footprints, categories, and support chains.

Fig.~\ref{fig:text-mask-pipeline} follows this representation through three
stages. Structure construction builds the room shell and openings. Placement
grounds each category to a procedural factory, sizes the asset to its authored
footprint, and instantiates it under orientation, containment, and support
constraints. Completion may add compatible content that the input did not
specify, but never replaces a prescribed object. Geometry validation,
materials, and lighting then finish the seeded scene that the 4D backbone
consumes.

\textbf{Real-to-sim compilation from a single photograph.}
A photograph is the most constrained input: the compiled scene has to
approximate one particular room, not just a plausible one.
Fig.~\ref{fig:r2s-pipeline} traces the evidence from a single RGB image to an
editable procedural scene. Monocular metric depth recovers camera-relative
room geometry~\cite{da3}, open-vocabulary segmentation finds objects and
openings~\cite{sam3}, object-centric reconstruction lifts instances to posed
3D proxies~\cite{sam3d}, and a multimodal agent maps each observed instance to
a procedural factory. Reconstruction thus supplies image-supported pose and
extent, and grounding supplies assets that can actually be built.

Metric and semantic reasoning combine this evidence into a route-specific
\emph{PerceivedScene}. Reconstructed proxies are anchored to the metric depth
and stored as metric 3D boxes with top-down footprints in the room frame;
category-level size priors correct what monocular scale bias remains. Support
relations and facing directions are estimated from geometry and image evidence
together, and the observed room boundary and openings become walls, windows,
and doors. The PerceivedScene also keeps the observed camera, object heights,
and appearance cues. A validated export maps its rooms, openings, objects, and
support relations onto the shared procedural backend and carries the
perception evidence along as build provenance.

Realization then follows the shared backend, with two additions that
faithfulness requires. First, geometry is settled conservatively: assets are
instantiated to their observed footprints while keeping their native shape,
placed through their support relations, and separated by mesh-level collision
resolution, so the recovered arrangement still holds once real procedural
geometry is in place. Second, appearance is aligned only after geometry is
frozen: for each room surface and object, an agent picks one of the native
material implementations, and albedo and exposure are calibrated against the
source view. Because alignment only ever chooses among procedural materials,
the compiled scene stays fully parametric---every wall, asset, and material
can still be edited---while its rendering approaches the photograph.
Throughout the compiler, agents choose only among bounded, schema-validated
alternatives, metric quantities are computed deterministically, and each stage
writes an immutable artifact, so a compiled scene can be audited and
regenerated one stage at a time.

\textbf{Camera-conditioned outdoor finalization.}
Indoor and prebuilt scenes expose their final geometry right away. Native
outdoor scenes do not: Infinigen materializes fine terrain and population
around the camera views, but a useful camera depends on where the actors can
walk. We break the circle with a coarse-to-final loop. A coarse proxy first
supports provisional actor routes and a provisional camera; their view frusta
then anchor the native population, fine terrain, ground cover, and
camera-local creatures. Once that detail exists, a new Stage is extracted and
both actors and camera are replanned on it.

The provisional routes are only priors. Final motion may bend around newly
realized terrain or vegetation, but it must stay collision-free within an
actor-sized connected component. The camera is repaired against the final
geometry in the same way and must stay inside the region that received native
detail. If no actor route satisfies the intent constraints, planning relaxes
them step by step and records the downgrade; if camera repair keeps failing, a
fresh whole-trajectory plan is made on the same scene. A different scene seed
is tried only before the scene checkpoint, when the first candidate offers no
actor-scale physical support. After finalization begins, every failure is
handled within the same scene.

\subsection{4D Animation on a Unified Stage}\label{sec:m-4d}

\textbf{Stage extraction.}
Animation starts by distilling the evaluated scene geometry into a
\emph{Stage}: a metric support grid with ground height and walkability, an
object inventory, and geometry for collision and visibility queries. One
representation serves indoor floors and outdoor terrain alike. A cell belongs
to the common walkable set $\mathcal{W}$ when it has stable, sufficiently
level support, is not deeply submerged, and has no solid geometry in the
vertical column an actor's body would occupy. Deformable ground cover such as
grass is excluded from collision on semantic grounds; rocks, trunks, and
structural objects stay obstacles.

For actor $i$ with measured body radius $r_i$, planning uses
$\mathcal{F}_i=\operatorname{Erode}(\mathcal{W},r_i)$. A scene is rejected if
it has no actor-scale physical support. Standable-area and corridor-length
targets are reported as diagnostics, and navigation additionally requires a
non-trivial route for every moving actor. Because the Stage also keeps terrain
height and the evaluated scene geometry, actor grounding, collision tests,
camera validation, and downstream task generation all refer to the same space.

\textbf{4D agent assets.}
Dynamic actors are characters with explicit geometry and time-varying
animation. A language model maps each requested action to the closest motion
in a curated library of motion-captured humanoid clips. Instead of normalizing
every character to a nominal humanoid, we measure each one after import:
stature, chest height, collision radius, visual envelope, sole offset, and
natural ground speed from root displacement. These numbers drive Stage
clearance, route erosion, inter-agent separation, terrain grounding, playback
rate, and camera framing; a clip with negligible root displacement is treated
as in-place.

\textbf{Collision-aware trajectory planning.}
Actors are placed and routed on the Stage through a layered feasible region.
The physically walkable set is the hard outer bound; optional preference
layers (currently a named-room prior) can only shrink it; and the result is
intersected with $\mathcal{F}_i$. Goal-directed actors resolve phrases such
as ``from the door to the sofa'' against the Stage inventory and plan with
A$^{\ast}$ between reachable anchors. Actors with no explicit goal get a long
corridor inside their connected free-space component. Outdoor proxy routes
contribute preferred endpoints or loop anchors, but the final geometry has the
last word and may force a local detour. Simplification and smoothing are
accepted only if every segment stays inside $\mathcal{F}_i$; the path is then
traversed at the clip's measured speed.

\textbf{Multi-agent coordination and baking.}
With several actors, reactive separation coordinates their motion while
keeping the hard feasible region and the planned endpoints. Stationary actors
are placed off the walking lines and treated as obstacles; narrow corridors
force single-file order. The tracks are then baked: root displacement removed,
animation cycles repeated as needed, heading aligned with the direction of
motion under a bounded turn rate, and feet grounded on the Stage height field.

\subsection{Geometry-Valid Observation Synthesis}\label{sec:m-camera}

A 4D scene is only useful if what the camera sees agrees with the world that
was synthesized. \sysname{} therefore keeps camera \emph{intent} separate from
geometric \emph{feasibility}. Intent is a timeline of nine composable
primitives---orbit, turn, push, pull, lift, lower, zoom, swoop, and static
hold---that can run in sequence or at the same time and are interpreted in
target-relative spherical coordinates. Framing distance follows from the
subject's extent and the field of view; it is not tuned per scene.

Feasibility and visibility are handled differently. There are two planning
modes, and both keep the camera inside the Stage's valid domain and outside
every measured actor body. The geometry-exact mode also requires the camera
origin to stay above support and outside solid scene geometry; invalid poses
are projected to nearby feasible ones, after which position, aim, field of
view, full-body framing, and reference-relative temporal correction are
checked again. The default fast mode instead evaluates 11 deterministic
whole-trajectory transforms. Terrain is a hard constraint in this mode, as is
any penetration whose owner (the scene object the penetrated surface belongs
to) cannot be identified as stable and safe to cut away. A confirmed
penetration---or an unresolved narrow-phase case backed by the same proof of a
stable, non-ground owner---may be handled by a frame-bounded render cutaway
instead of a frame-local camera detour.

Occlusion by the environment is a soft cost. The search prefers clear views
but does not detour around every foreground branch or piece of furniture; such
occluders may stay in frame, defocused around the tracked subject. In fast
planning, the fraction of frames in which every target is fully framed with an
unoccluded chest sample must meet the requested threshold, which is never
allowed below $80\%$. Exact planning runs a terminal audit on the full
render-visible geometry and, by default, requires each target to stay present
and fully framed for at least $90\%$ of the interval after it is first
revealed. Brief natural occlusion passes; sustained loss of the target does
not. The same machinery handles moving third-person views, static views of
objects, and transitions from a scene's native establishing camera.

\subsection{Physics-Ready Export to Simulation}\label{sec:m-export}

The exporter writes a finished 4D scene to an OpenUSD stage for Isaac Sim. The
whole scene is exported once and checked by content---mesh and prim
counts---rather than by exit status. Light sources whose parameters do not
survive the transfer are repaired; outdoor scenes also get a sun light and
smoothed terrain normals.

Physical properties come from the generative representation itself. Every
surface carries a procedural material, so material identity maps to density
and friction, and object-level values are area-weighted aggregates over
material regions. Every mesh gets a static collider; objects marked movable
also become rigid bodies with convex-hull collision and density-derived mass.
No separate physics labeling is needed: the annotations are read off the same
procedural parameters that produce the appearance.

Animated humans enter the simulator as kinematic solid obstacles. Each actor
is a capsule of its measured radius and height that follows the generated
track, sampled at the simulator's playback rate. Kinematic obstacles push and
block but are not pushed, which is the role pedestrians play in navigation.
The capsule is the collision surrogate on the simulator side; the skinned
character remains a render-visible animation asset. Before any benchmark task
is generated, each exported scene is verified headlessly in Isaac Sim: the
stage must load, dynamic objects must settle under gravity, and each actor
must reproduce its planned displacement during playback.

\section{Experiments}\label{sec:experiments}

\subsection{Qualitative Scene Synthesis}\label{sec:exp-synthesis}

We first look at the range and editability of the scene routes qualitatively.
Everything shown is rendered directly from the synthesized procedural scenes;
the comparisons illustrate how the system behaves and are not a metric
ranking.

\begin{figure*}[!t]
  \centering
  \includegraphics[width=\textwidth]{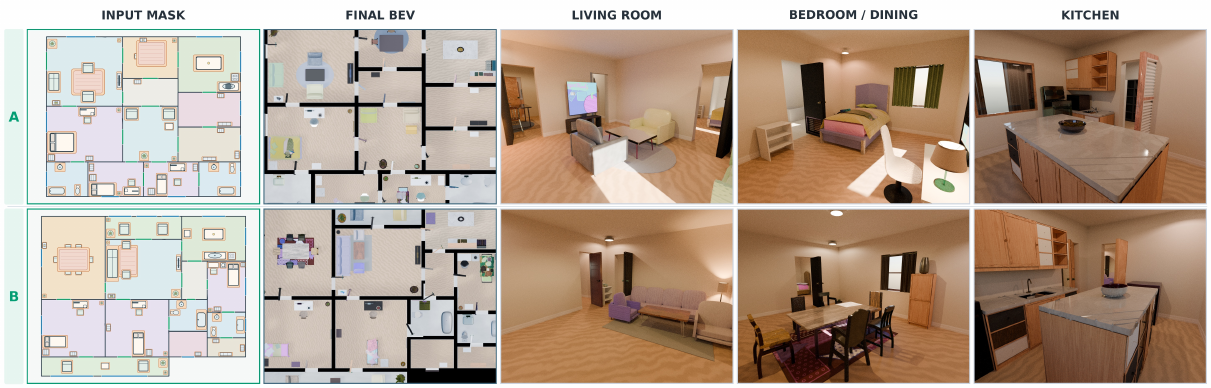}
  \caption{Layout-conditioned synthesis for two blueprint masks (A and B).
  Each row shows, left to right, the authored input, the final procedural
  bird's-eye view (BEV), and selected living-room, bedroom/dining-room, and
  kitchen renders. The multi-room organization and the authored furniture
  arrangement are preserved, and the 2D constraints become editable procedural
  geometry with room-specific materials and lighting.}
  \label{fig:qual-textmask}
\end{figure*}

\textbf{Layout-conditioned synthesis.}
Fig.~\ref{fig:qual-textmask} shows two multi-room homes compiled from
blueprint masks. Each row pairs the authored input with the final top-down
scene and selected living-room, bedroom/dining-room, and kitchen views. Room
topology, openings, and furniture arrangement are still recognizable after
realization, while the procedural factories supply 3D geometry, materials, and
lighting. Since the outputs are native procedural scenes, individual assets
and room surfaces can be edited afterwards.

\begin{figure*}[!t]
  \centering
  \includegraphics[width=0.99\textwidth]{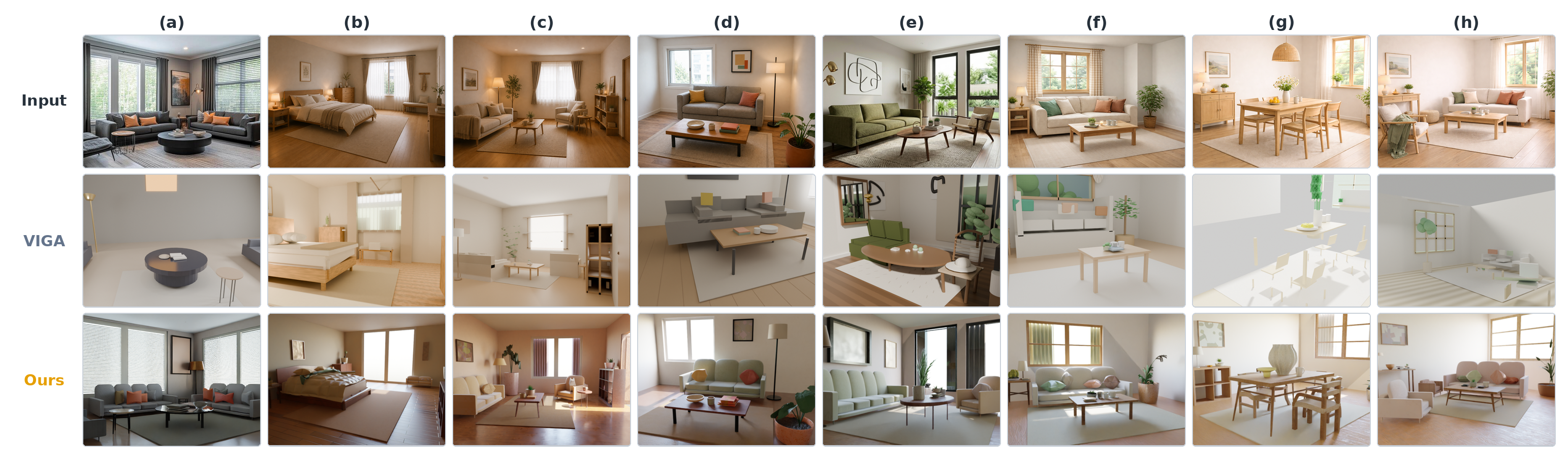}
  \caption{Qualitative comparison with VIGA on eight shared single-image
  inputs. Rows show the input photograph, the VIGA reconstruction, and our
  appearance-aligned procedural result under a common 4:3 viewport. Across
  these examples our compiler more consistently recovers explicit room
  surfaces, openings, secondary furniture, and separately editable object
  instances while keeping the dominant layout and appearance. The comparison
  is qualitative; we do not infer a metric ranking from it.}
  \label{fig:qual-r2s-viga}
\end{figure*}

\textbf{Real-to-sim compilation.}
Fig.~\ref{fig:qual-r2s-viga} compares input photographs, VIGA
reconstructions~\cite{viga}, and our compiler's output on eight shared
single-image inputs. In ours, room geometry, furniture layout, and support
relations follow the photograph, and the appearance-alignment stage matches
wall, floor, and furniture materials to the source view. The result is not a
textured reconstruction; it is a procedural scene that the downstream 4D
pipeline can animate, relight, and edit. On these examples VIGA recovers
several visually dominant objects but often drops openings or secondary
furniture and distorts their relative placement, whereas our compiler keeps
more of the room envelope, object inventory, and support structure as
separately editable procedural instances.

\begin{figure*}[!t]
  \centering
  \includegraphics[width=0.99\textwidth]{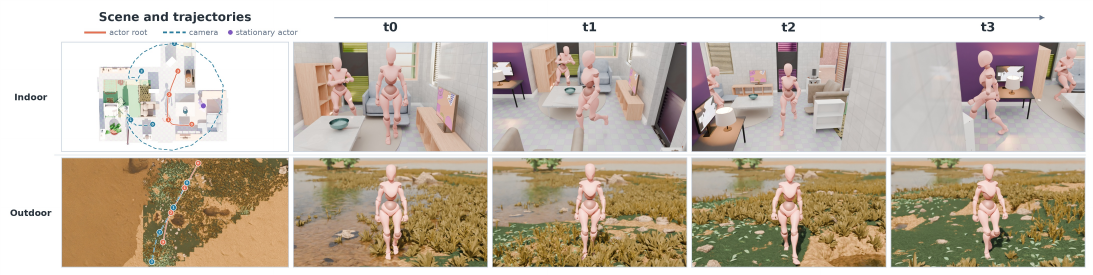}
  \caption{Synthesized 4D environments. Each row pairs an overhead rendering
  of the final scene with four frames sampled at the numbered times. Solid
  orange and dashed blue curves are the measured actor-root and solved camera
  trajectories; the purple marker in the indoor scene is the in-place dancing
  actor. Environment, actors, and camera are baked into one editable 4D
  scene.}
  \label{fig:qual-4d}
\end{figure*}

\textbf{4D environments.}
Fig.~\ref{fig:qual-4d} shows frames from synthesized 4D scenes. Moving actors
follow collision-free trajectories indoors and across native outdoor terrain,
in-place actions stay at their validated placements, and the camera keeps its
framing against the final geometry. The physics-enabled scenes behind the
navigation benchmark come from this same pipeline, with no per-scene mesh
authoring.

\subsection{Benchmark Construction}\label{sec:benchmark}
We build \textbf{4DSynth-Nav} from 122 physics-enabled indoor scenes
generated with Infinigen Indoors~\cite{infinigenindoors} and simulated in
NVIDIA Isaac Sim~4.5.0. Each scene contains procedural household objects and
one or more animated characters following baked trajectories (solo run, two
runners, run--dance, or run--jump). The characters are dynamic obstacles; they
do not react to the agent.

\noindent\textbf{Automated task generation.}
A three-phase pipeline produces validated navigation and pick-and-place tasks
with no per-task manual annotation:
\begin{enumerate}
    \item \textbf{Scene probing}: Each USD stage is loaded in Isaac Sim. The
    active geometry yields object prim paths,
    world-space bounding boxes, floor heights, and support relations (by
    bounding-box stacking). A PhysX flood fill on a 0.25\,m grid, with sphere
    sweeps at $z{=}0.5$\,m and $z{=}1.0$\,m for a 0.40\,m-radius agent, gives
    per-object reachability from the floor centroid of the primary room.

    \item \textbf{Task generation}: This pure-Python phase selects targets
    under five invariants: (i)~pickups sealed in closed cabinets are rejected;
    (ii)~semantic classes that are unique in the room are preferred, so that
    no same-class distractors need to be deactivated; (iii)~no phase target
    may share the semantic class of a pickup's support furniture; (iv)~floor
    pickups are capped at 30\%, since they force a downward camera tilt that
    hurts vision-language model (VLM) performance; and (v)~targets must lie
    within validated reachability radii (1.0\,m for pickups, 1.5\,m for
    destinations). Any remaining same-class non-targets, with the clutter
    resting on them, go into a per-task deactivation list computed against
    the realized geometry.

    \item \textbf{Spawn validation}: Each candidate is loaded in Isaac Sim,
    and the spawn must (a)~lie inside the exact concave room boundary
    extracted from single-occurrence USD mesh edges; (b)~pass eight-directional
    PhysX sphere sweeps at two heights; (c)~have the phase-1 target inside the
    initial frustum ($\pm45^{\circ}$ horizontally, $\pm29^{\circ}$ vertically)
    for L2 tasks and outside it for L1/L3 (tiers are defined in
    Section~\ref{sec:dataset}); (d)~pass a 1.2\,m forward-clearance raycast;
    and (e)~keep flood-fill reachability to the phase-1 target. 
\end{enumerate}
\subsection{Dataset Statistics}\label{sec:dataset}
\begin{table}[t]
\centering
\caption{4DSynth-Nav benchmark composition.}
\label{tab:dataset}
\scriptsize
\setlength{\tabcolsep}{1.5pt}
\begin{tabular}{@{}ll@{}}
\toprule
\textbf{Property} & \textbf{Value} \\
\midrule
Total tasks & 333 (L1: 113, L2: 110, L3: 110) \\
Unique scenes & 113 (from 122 physics-enabled Infinigen Indoors scenes) \\
Room types & living room, bedroom, kitchen, dining room, bathroom \\
Task types & navigate (113), two-waypoint nav (130), pick-and-place (90) \\
Pickup objects & 11 categories (book, cup, bottle, bowl, vase, \ldots) \\
Destination objects & 11 categories (shelf, desk, counter, toilet, \ldots) \\
Dynamic obstacles & solo run (37\%), two runners (12\%), run--dance/jump (51\%) \\
\bottomrule
\end{tabular}
\end{table}
The benchmark has three tiers (L1--L3) along two controlled axes:
\emph{number of subtasks} and \emph{initial target visibility}. The 
combination of one phase with the target already in view- is left out because
it exercises neither difficulty factor.
\begin{itemize}
    \item \textbf{L1} (one phase, initially hidden): the agent must turn or
    explore to find a single target outside its initial camera frustum.
    \item \textbf{L2} (two phases, initially visible): navigate--then--navigate
    or pick--then--deliver, with the phase-1 target inside the initial frustum.
    About 60\% of phase-1 targets are navigation waypoints and 40\% are
    pickups.
    \item \textbf{L3} (two phases, initially hidden): the L2 task structure
    with the blind start of L1.
\end{itemize}
\subsection{Experimental Setup}\label{sec:setup}
\noindent\textbf{Agent architecture.}
The agent receives up to three recent first-person images
(960${\times}$540, path-traced), the task instruction in natural language, and
feedback on which actions were executed and whether motion was blocked. At
each decision step it emits a plan of at most five actions from
\texttt{MOVE\_FORWARD} (0.25\,m), \texttt{TURN\_LEFT/RIGHT} ($15^{\circ}$),
\texttt{TILT\_UP/DOWN} ($5^{\circ}$), \texttt{PAUSE}, \texttt{PICK\_UP},
\texttt{PUT\_DOWN}, and \texttt{DONE}. Plans run sequentially and stop on
collision or phase completion. An episode ends after 150 actions or 50 VLM
calls, whichever comes first.

\noindent\textbf{Models.}
We evaluate \textbf{Qwen3-VL-30B-A3B-Thinking}~\cite{qwen3vl}, a
mixture-of-experts vision-language model with 30B total and 3B active
parameters, served with vLLM, and the proprietary
\textbf{Gemini~3.1~Pro}~\cite{gemini31pro}, on the same 333 tasks. The two
differ in architecture, training data, and serving stack, so we treat them as
diagnostic probes rather than as a controlled comparison of navigation agents.

\noindent\textbf{Metrics.}
We report \emph{Success Rate} (SR), the fraction of episodes in which all
phases are completed; \emph{Subtask Progress} (SP), the fraction of phases
completed; \emph{Goal Distance} (GD), the Euclidean distance to the current
target at episode end; \emph{Collisions}, the mean number of recorded obstacle
contacts per episode; and \emph{Steps}, the mean number of steps per episode.
\subsection{Results}\label{sec:results}
\begin{table}[t]
\centering
\caption{4DSynth-Nav results. SR and SP in \%; GD in meters; Coll.\ and Steps
are per-episode means.}
\label{tab:results}
\normalsize
\setlength{\tabcolsep}{4pt}
\begin{tabular}{@{}llccccc@{}}
\toprule
\textbf{Model} & \textbf{Level} & \textbf{SR} & \textbf{SP} & \textbf{GD} & \textbf{Coll.} & \textbf{Steps} \\
\midrule
\multirow{4}{*}{Qwen3-VL-30B}
 & L1\, & 16.8 & 16.8 & 4.74 & 43.2 & 120.2 \\
 & L2\, & 18.2 & 35.5 & 3.84 & 36.6 & 103.8 \\
 & L3\, &  4.5 & 15.0 & 4.22 & 37.8 & 117.2 \\
 & All\, & 13.2 & 22.4 & 4.27 & 39.2 & 113.8 \\
\midrule
\multirow{4}{*}{Gemini 3.1 Pro}
 & L1\, & 57.5 & 57.5 & 2.41 & 17.0 & 79.5 \\
 & L2\, & 23.6 & 36.4 & 2.42 & 20.2 & 114.0 \\
 & L3\, & 18.2 & 27.7 & 3.23 & 19.9 & 120.7 \\
 & All\, & 33.3 & 40.7 & 2.68 & 19.0 & 104.5 \\
\bottomrule
\end{tabular}
\end{table}
\noindent\textbf{Our benchmark can differentiate model capacity, and there is still room for the models to solve the benchmark.}
As shown in Table~\ref{tab:results}, Qwen3-VL-30B completes 13.2\% of tasks and 22.4\% of subtasks. Gemini~3.1~Pro
raises SR to 33.3\% and SP to 40.7\% and cuts the final goal distance from
4.27\,m to 2.68\,m. This is a large gain in both completion and geometric
execution, but it still fails two of every three tasks. Neither model is
close to solving dynamic embodied navigation in these scenes.

\noindent\textbf{Gemini represents better navigation capability  on blind starts.}
On L1, where the target starts outside the agent's field of view, Gemini
reaches 57.5\% SR against 16.8\% for Qwen, in fewer steps (79.5 vs.\ 120.2)
and ending closer to the goal (2.41\,m vs.\ 4.74\,m). On the two-phase tiers
the gain is smaller on L2 but still large on L3, where SR goes from 4.5\% to
18.2\%. Task types differ across tiers and the models differ in training, so
we read these numbers as diagnostics, not as a controlled ablation.

\noindent\textbf{Both models get partway and then stall.}
Subtask progress runs well ahead of success: for Qwen, 35.5\% SP against
18.2\% SR on L2 and 15.0\% against 4.5\% on L3; for Gemini, 36.4\% against
23.6\% on L2 and 27.7\% against 18.2\% on L3. The pattern fits compounding
errors in target search, spatial memory, manipulation, and re-navigation after
a phase transition.

\subsection{Trajectory Example}
\begin{figure}[t]
    \centering
    \includegraphics[width=\linewidth]{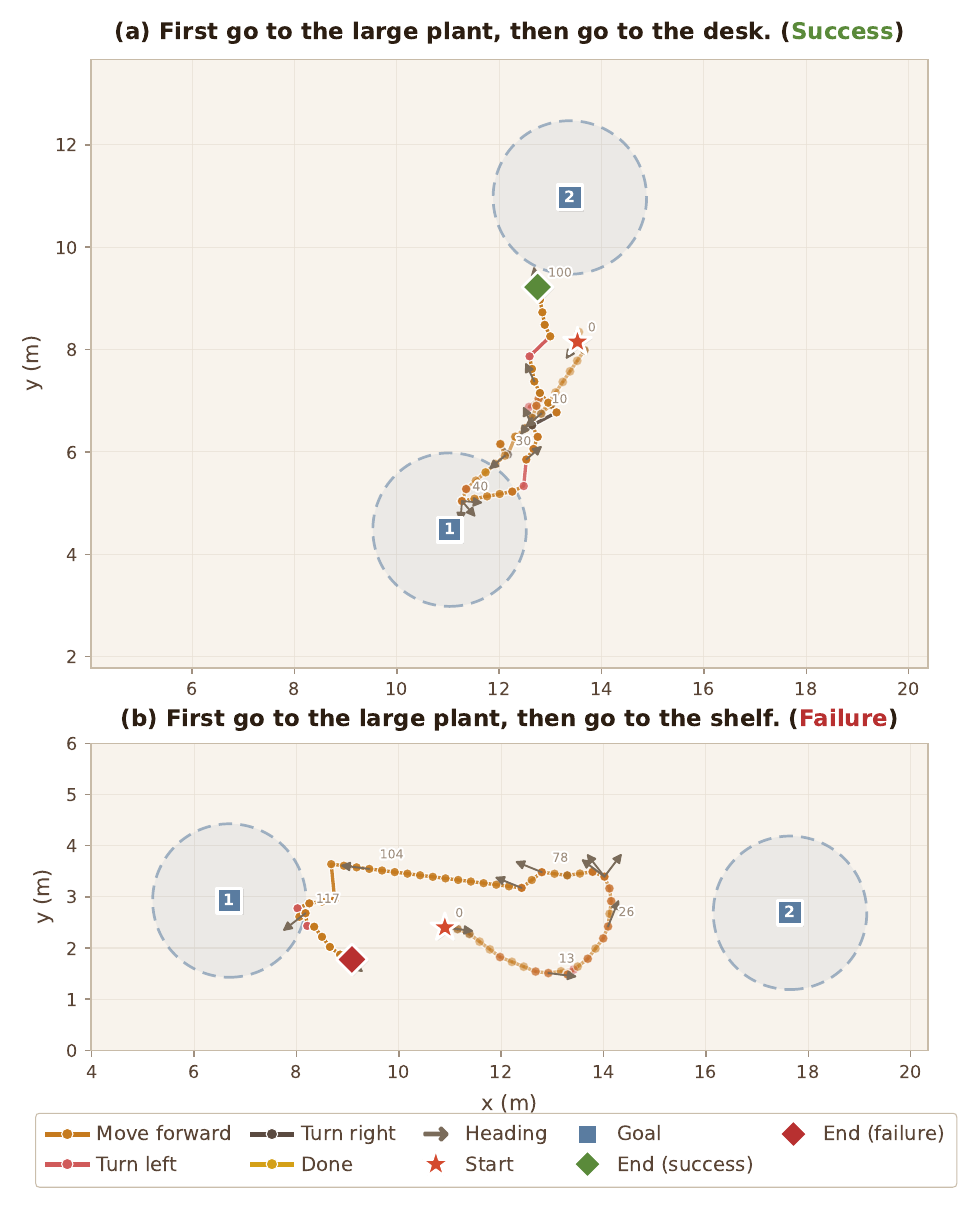}
    \caption{Trajectory diagnostics for Qwen3-VL-30B on two-goal navigation.
    \textbf{(a)}~L2 success: both goals reached in 101 steps, ending 1.3\,m
    from Goal~2. \textbf{(b)}~L3 failure: Goal~1 reached, then the agent loops
    away from Goal~2 and ends 8.0\,m from it (SP$=0.5$).}
    \label{fig:traj-compare}
\end{figure}

Fig.~\ref{fig:traj-compare} contrasts a successful and a failed two-goal
episode, both run by Qwen3-VL-30B. In the L2 success
(Fig.~\ref{fig:traj-compare}(a)), the agent closes in on Goal~1 (the large
plant), enters its acceptance region within 40~steps, turns toward Goal~2
(the desk), and finishes both phases in 101~steps, 1.3\,m from the desk.

The L3 failure (Fig.~\ref{fig:traj-compare}(b)) has the same two-goal
structure, but the phase-1 target starts outside the camera frustum. The agent
still reaches Goal~1, by step~104 (SP$=0.5$), and the harness then advances
the objective prompt to \emph{``Current objective: go to shelf \ldots\
Progress: step 2/2.''} Even with that cue, the agent traces a wide loop away
from the shelf and ends 8.0\,m from it after 133~steps. The two episodes demonstrate that the model is capable of solving multi-stage goals, but misjudgment of task completion status can still happen.

% \subsection{Limitations}\label{sec:limitations}
% Our evidence has four important boundaries. First, scene synthesis is evaluated
% primarily through qualitative examples; the eight VIGA comparisons do not
% establish geometric or photometric superiority. Second, monocular real-to-sim
% is inherently ambiguous under occlusion and substitutes procedural assets for
% observed instances, so it produces editable scene analogues rather than exact
% digital twins. Third, benchmark characters are kinematic obstacles with a
% small set of scripted motions and cannot react socially or physically to the
% agent. Finally, the benchmark evaluation uses two VLMs and 113 contributing
% indoor scenes without repeated rollouts. The comparison is descriptive rather
% than controlled because the open and proprietary models differ in architecture,
% training data, and serving stack. Additional models, repeated trials,
% route-level quantitative metrics, and broader human motions are needed before
% drawing general model or scalability conclusions.

\section{Conclusion}
We presented \sysname{}, a procedural pipeline that builds editable 4D
environments from natural-language descriptions, blueprint masks, and single
photographs. Native and compiled scene routes meet in a shared Stage of
finished geometry, walkability, and scene content, and a WorldState ties that
Stage to the actor and camera trajectories and to the provenance of every
intermediate result. Rendering, physics simulation, and task generation all
read from these same representations, and because the environments keep their
procedural structure, they can be regenerated, edited, animated, and exported
without any per-scene mesh work.

We demonstrated the pipeline with 4DSynth-Nav, 333 navigation and
pick-and-place tasks with static and dynamic obstacles. For both Qwen3-VL and
Gemini~3.1~Pro, the hardest tier is the one that combines a blind start with a
two-stage objective, and partial progress, collision counts, and looping
trajectories reveal failures that success rate alone hides. Next we plan to
quantify route-level fidelity and controllability, add responsive humans and
articulated objects, and test a broader set of embodied agents.

\end{document}